\documentclass[sigconf,10pt]{acmart}
\usepackage[font=small,labelfont=bf]{caption}

\AtBeginDocument{%
  \providecommand\BibTeX{{%
    \normalfont B\kern-0.5em{\scshape i\kern-0.25em b}\kern-0.8em\TeX}}}

 \setcopyright{none}
 \renewcommand\footnotetextcopyrightpermission[1]{}%

\usepackage{hyperref}

\usepackage{microtype}
\usepackage{graphicx}
\usepackage{multirow}
\usepackage{makecell}
\usepackage{booktabs} %
\usepackage{paralist}
\usepackage{xspace}
\usepackage{float}
\DeclareUnicodeCharacter{211D}{\mathbb{∈}}

\usepackage[utf8]{inputenc} %
\usepackage[T1]{fontenc}    %
\usepackage{url}            %
\usepackage{booktabs}       %
\usepackage{amsfonts}       %
\usepackage{amsmath}       %
\usepackage{nicefrac}       %
\usepackage{microtype}      %
\usepackage{algorithmicx}
\usepackage{algorithm}%
\usepackage{algpseudocode}%

\usepackage{mwe}    %
\usepackage{float}
\usepackage{lineno}
\usepackage{caption}
\usepackage[skip=0cm,list=true]{subcaption}

\usepackage{xcolor}
\definecolor{codegreen}{rgb}{0,0.6,0}
\definecolor{codegray}{rgb}{0.5,0.5,0.5}
\definecolor{codepurple}{rgb}{0.58,0,0.82}
\definecolor{backcolour}{rgb}{0.95,0.95,0.92}

\newcommand{\smartparagraph}[1]{\noindent{\bf #1}\ }
\newcommand{\scheme}{\textit{EAFL}\xspace}

\newcommand{\K}{\textbf{K}\xspace}
\newcommand{\B}{\textbf{B}\xspace}
\newcommand{\E}{\textbf{E}\xspace}
\newcommand{\N}{\textbf{N}\xspace}

\usepackage[capitalise]{cleveref}
\usepackage{paralist}
\usepackage{enumitem}[noitemsep,topsep=0pt,leftmargin=10pt]
\setitemize{noitemsep,topsep=0pt,parsep=0pt,partopsep=0pt}
\setenumerate{noitemsep,topsep=0pt,parsep=0pt,partopsep=0pt}

\newif\ifsubmission
\submissionfalse

\ifsubmission
\newcommand{\ahmed}[1]{}
\newcommand{\amna}[1]{}
\else
\newcommand{\ahmed}[1]{\textit{\textcolor{red}{{Ahmed:}#1}}}
\newcommand{\amna}[1]{\textit{\textcolor{blue}{[amna]: #1}}} 
\fi

\makeatletter
\def\runningfoot{\def\@runningfoot{}}
\def\firstfoot{\def\@firstfoot{}}
\makeatother

\begin{document}

\title{Towards Energy-Aware Federated Learning on Battery-Powered Clients}

\author{Amna Arouj}
\affiliation{\institution{Queen Mary University of London}\country{United Kingdom}}
\authornote{Work done during an internship at Queen Mary University of London.}
\author{Ahmed M. Abdelmoniem}
\affiliation{\institution{Queen Mary University of London}\country{United Kingdom}}
\authornote{Corresponding author (\url{ahmed.sayed@qmul.ac.uk}), also with Assiut University, Egypt}

\begin{abstract}
Federated learning (FL) is a newly emerged branch of AI that facilitates edge devices to collaboratively train a global machine learning model without centralizing data and with privacy by default. However, despite the remarkable advancement, this paradigm comes with various challenges. Specifically, in large-scale deployments, client heterogeneity is the norm which impacts training quality such as accuracy, fairness, and time. Moreover, energy consumption across these battery-constrained devices is largely unexplored and a limitation for wide-adoption of FL. To address this issue, we develop EAFL, an energy-aware FL selection method that considers energy consumption to maximize the participation of heterogeneous target devices. \scheme is a power-aware training algorithm that cherry-picks clients with higher battery levels in conjunction with its ability to maximize the system efficiency.
Our design jointly minimizes the time-to-accuracy and maximizes the remaining on-device battery levels. \scheme improves the testing model accuracy by up to 85\% and decreases the drop-out of clients by up to 2.45$\times$.\footnote{Accepted to appear in FedEgde Workshop of ACM MobiCom 2022\\Code and scripts are available at \url{https://github.com/SAYED-Sys-Lab/EAFL}} 

\end{abstract}
\keywords{Federated Learning, Heterogeneity, Performance, Fairness}

\maketitle
\section{Introduction}
\label{sec:intro}

Due to the advancements in technology and the wireless industry's growth, a wealth of data is born at the edge every day. We are connecting more devices and crunching data faster than ever before and by 2025 the number of smartphones and wearable devices will reach 7.33 and 1 billion, respectively~\cite{the-wave-of-wearables}. 
To draw useful information from this geographically distributed data, a prominent paradigm known as federated learning (FL) has emerged that allows the end devices to learn a shared ML model while preserving the user's data privacy. Ideally, a cluster of users run stochastic gradient descent (SGD) locally and aggregate their updated models via the server to obtain a new global model.

Lately, the advancements in the powerful mobile System-on-Chips (SoCs) further foster the adoption of FL for collaborative learning on  edge devices such as smartphones, smart-wear, IoT devices, etc. These devices are now equipped with high-performance central processing units (CPUs) and graphics processing units (GPUs) to operate intensive computations for AI/ML-based applications. Due to the technical abilities and implementation ease, FL has recently seen wins in several applications like power keyboard predictions(Gboard), vocal/face classifiers (Face ID and Siri), virtual assistants, smart cities and augmented reality~\cite{autofl}.

Although on-device inference comes with promising deployment in exciting applications, improved latency, work offline, privacy advantages, and better battery life, however, implementing FL on mobile devices brings severe challenges, of which energy consumption is a primary concern. Common characteristics of these devices are that they have limited energy, storage, and computing resources; thus, because of these constraints, optimizing the energy efficiency of the ML inference while fulfilling the Quality-of-Service (QoS) requirements is essential for these services.

Although existing FL solutions have shown significant progress in overcoming the challenges encountered in the FL design space, most of them, however, given a pool of participants, focus either on optimizing \textit{statistical model efficiency} (i.e., improving training accuracy with lesser training rounds) \cite{Li2020FedProx, domain}  or \textit{system efficiency} (i.e., shorter training rounds) \cite{mcmahan2017, distributed} while other mechanisms have been proposed to guarantee privacy and robustness \cite{Keith, clientlevel}. As adopted in Oort, the designer makes all efforts to optimize the system efficiency, ignoring the diversity of the clients’ data \cite{Oort-osdi21}. This level of unfairness results in a less robust model toward data heterogeneity and, therefore, results in 

Executing intensive on-device computation for long periods can quickly drain the battery or burn the device, leading to client dropouts \cite{FedScale}. Hence, we deem \textit{a power-aware FL training is an open problem}. In this work, we introduce \scheme, a novel, user-friendly training algorithm that picks high remaining power learners to increase the participation level in FL training and reduce the energy impact on user's devices. %
 Our results show that \scheme delivers significant accuracy benefits over the state-of-the-art while reducing drop-outs due to battery drainage, hence preserving the user experience. In this work, we make the following contributions:
\begin{enumerate}
    \item We consider a more practical FL scenario on battery-powered devices where heavy computations during FL training may lead to performance degradation while accounting for the heterogeneity in the system. %
    \item Based on studying various energy-consumption models of real mobile devices, we present a power-aware design that intelligently improves the FL performance in battery-powered scenarios through reduced client dropouts and increased participation levels.%
    \item We show, via experiments on real FL benchmark in battery-powered scenario, that \scheme produces models of high quality compared to state-of-art solutions.
\end{enumerate}

\section{Background and Motivation}
\label{sec:background}

\begin{figure}[!t]
    \centering
    \includegraphics[width=1\columnwidth]{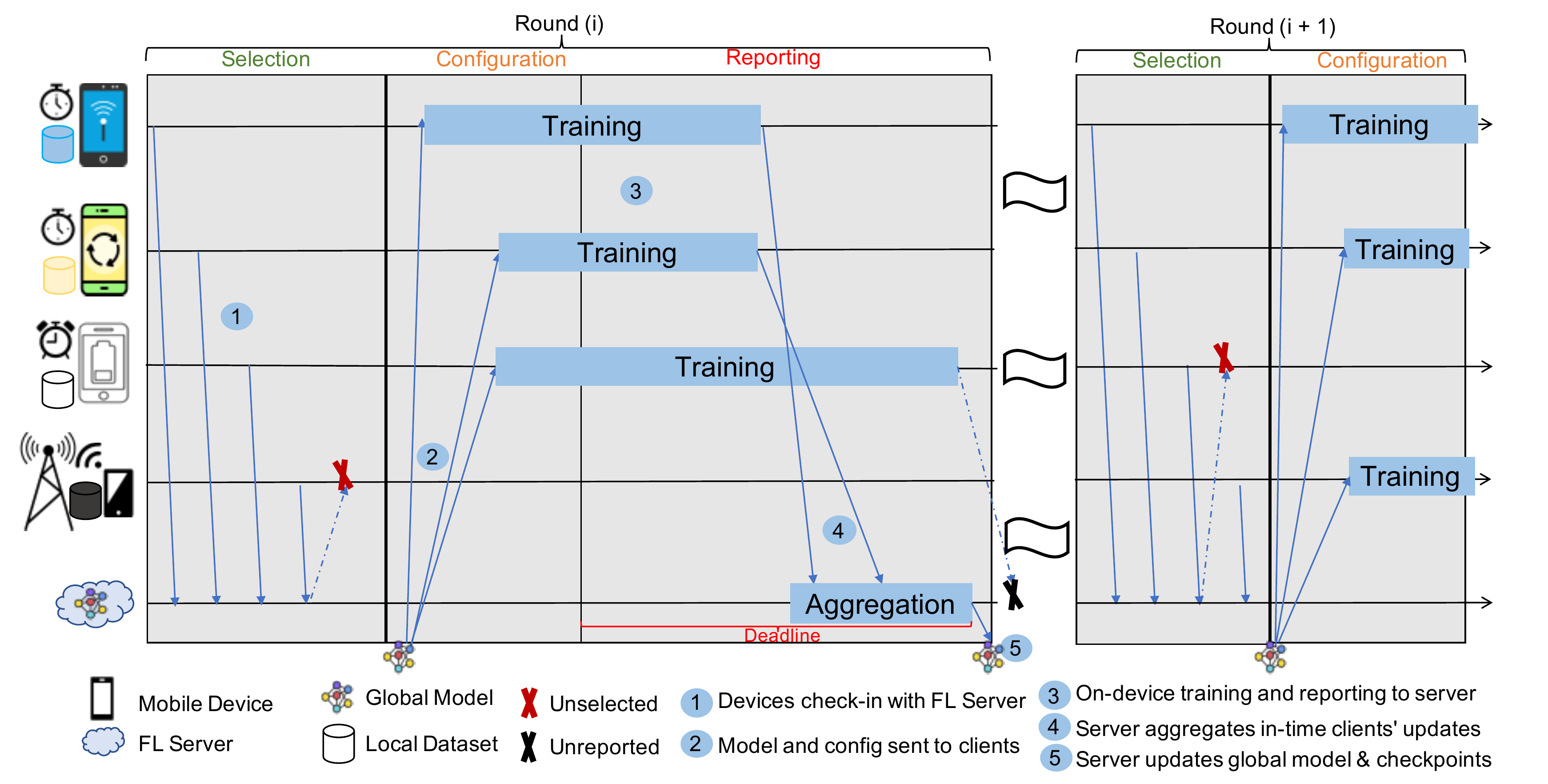}
    \caption{The various phases in training rounds of FL.}
    \label{fig:fedlearn}
\end{figure}

We start with a quick overview of the FL system design, followed by
highlighting the key limitations in the existing solutions that motivate our work.  
\subsection{Federated learning}
Federated learning, an emerging networking paradigm, runs machine learning on decentralized data. It takes advantage of the available computing resources across a massive pool of edge devices. More specifically, all clients collectively train on one global learning model and perform regular model parameter updates by continuous interactions. 

Two main entities are in the core of FL design ---\emph{Clients}, the data owners (e.g. smartphones, tablets, laptops, auto-vehicles, etc.) and --- \emph{Aggregator},  the model owner (e.g. server). Training of the global model takes place by specifying the number of epochs \E (rounds) until the model converges. Other parameters like minibatch size \B, and participants per round \K are also specified and are determined by the FL-based services~\cite{Bonawitz19,autofl}. As depicted in \cref{fig:fedlearn}, at the start of each training round, the aggregator selects \K participants among \N available devices depending on given criteria. \emph{(Step 1)} Server broadcasts the model's current version and other necessary hyper-parameters to the selected devices. \emph{(Step 2)} Each participant trains the model by performing \E local optimization steps with a batch size of \B. \emph{(Step 3)} Learners send their computed model updates back to the server.\emph{(Step 4)}The server collects the model updates from the participants for aggregation and checkpoints it. \emph{(Step 5)} The stages are repeated until the desired accuracy is achieved.

\subsection{Motivation}
In this part, we try to motivate our work.

\smartparagraph{FL considerations in mobile environment:} Existing FL solutions have simply failed to notice the consequences resulting in the interplay of client devices and training speed  (e.g., using significant local steps to save communication) \cite{mcmahan2017} leads to intensive on-device computation for extended periods, resulting in the sudden drain of the battery or even burning of the device hence leading to unavailability of the client. However, the possibility of redeeming these disadvantages by implementing a \emph{power-aware training algorithm} has been incredibly overlooked~\cite{FedScale}.

Furthermore, very less work is presented in the \emph{energy efficiency} optimization domain. Most prior work supposes that FL training is activated once smartphones are plugged into a power-point because of the significant energy consumption during FL training~\cite{C.Wang,save}. Unfortunately, this has restricted the practical implementation of FL, resulting in incorrect model accuracy and long convergence times. Energy-efficient federated learning can facilitate on-device training with better accuracy, model quality, and user experience. Through \scheme, we aim to present solutions to simultaneously achieve system and energy conservation, embracing the opportunities presented by earlier work.

\smartparagraph{System heterogeneity and client drop-outs:} Edge and IoT devices differ in underlying architectures (CPU, design, or memory). These devices function on different battery levels and use different communication mediums (e.g., Wi-Fi or 4G/5G). Hence, each device differs in computational storage and communication capacities. These differences in system configurations introduce problems to FL schemes especially when entry-level edge devices with low battery power and bandwidth are involved. This is because these devices are more susceptible to dropouts during training when they run out of battery at any time. 

Client drop-out is synonymous with the straggler issue in FL training, in which some clients delay the uploading of the local model~\cite{mcmahan2017}. Many solutions exist to mitigate the straggler issue~\cite{Xie2019,straggler}. A straggler-resilient design is proposed in \cite{straggler} that adaptively selects clients by incorporating the statistical characteristics of the client's data. However, client drop-out is a more severe case because dropout clients cannot upload their model in the current round and are likely to remain unavailable for some period of time; thus, existing schemes to mitigate the straggler issue do not work~\cite{friends}.

For the aforementioned reasons, system heterogeneity introduces unexplored challenges in FL. We introduce the notion of power-aware training which mitigates client drop-out and improves FL performance. Our design goal is to trade-off remaining power with time which reduces the clients' dropouts and hence improves the model quality.

\vspace{-0.5em}
\section{EAFL Design Overview}
\label{sec:methodology}

\begin{figure}[!t]
    \centering
    \includegraphics[width=0.9\columnwidth]{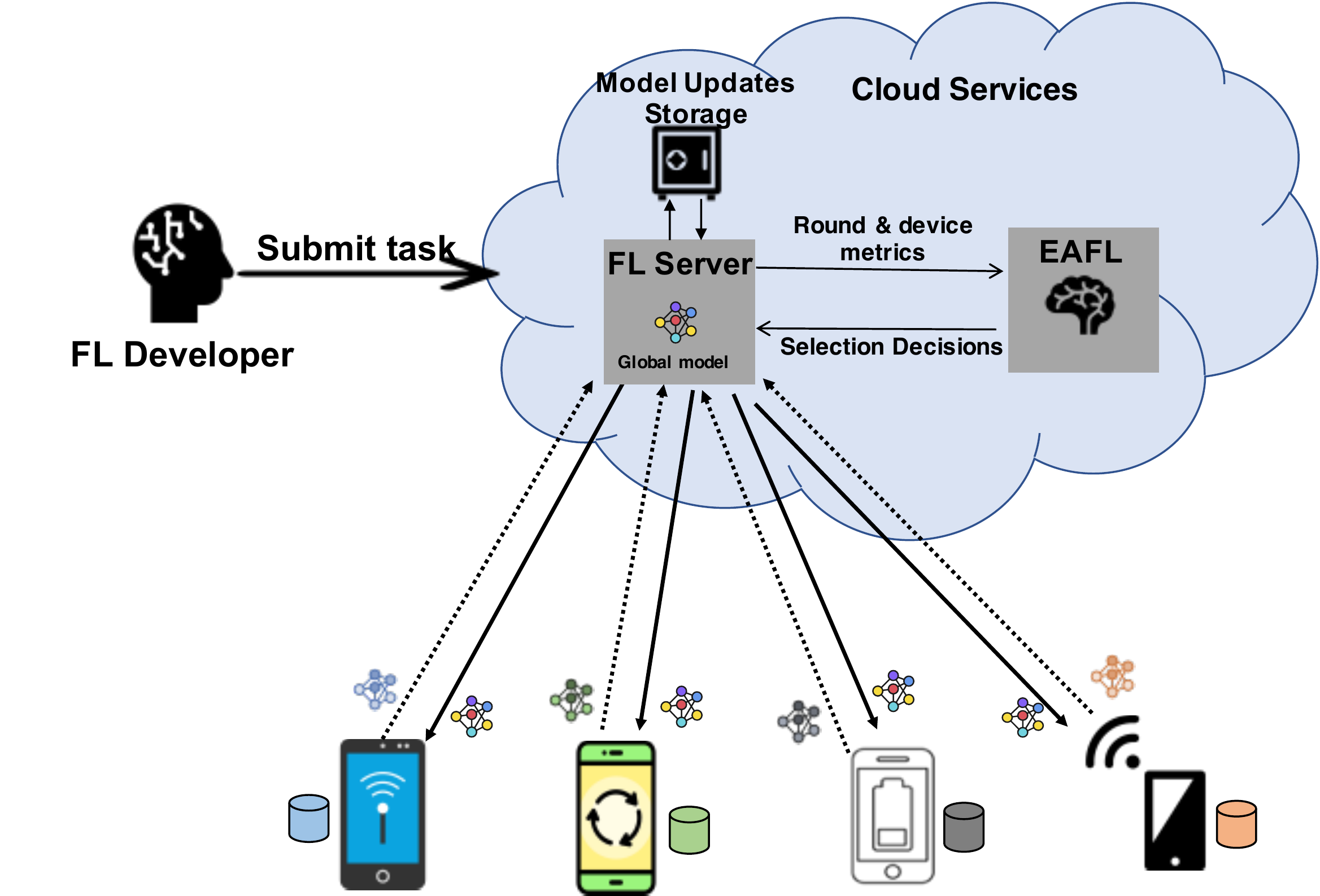}
    \caption{High-level Architectural Design of \scheme.}
    \label{fig:eafl}
\end{figure}

\scheme's goal is to jointly reconcile the demand for power-aware FL training and optimization of system/statistical efficiency over heterogeneous battery-powered edge devices. To illustrate and study the heterogeneity impact on
model quality and client drop-outs, we follow the experimental
design approach~\cite{the-design-of-experiments} and raise the following questions:

\begin{enumerate}
    \item To what extent do client dropouts due to battery-constraints impact  the convergence of FL?
    \item What is the trade-off between model quality and energy efficiency of the system?
\end{enumerate}

\subsection{Architecture}
\cref{fig:eafl} shows the high-level architecture of \scheme and the interactions between \scheme, the FL developer and server and the edge devices. The developer submits the task to the FL coordinator, and then the coordinator registers each client's profile (e.g., battery level, workload, RAM, etc.) and forwards the characteristics to the server running \scheme. Based on the feedback from the earlier rounds and client profiles, \scheme associates a utility with each client and selects the group of participants for the upcoming training round according to the selection mechanism. Next, the coordinator distributes the relevant model profiles to each selected participant, who computes results on their local data individually. The coordinator then collects the updates from the participants to aggregate. Furthermore, the steps from selection to aggregation are repeated until the desired accuracy is achieved.

In \scheme, the algorithm identifies and favours participants with higher utilities. The reward function calculates the utility of each client which consists of two parts. One part is a function for jointly measuring the system and statistical utility and the other part is a function of the remaining battery level of the client's device. We test our scheme under different scenarios by giving different weights to each function in the utility definition. We elaborate more on client utility in \cref{sec:training}.  In our scheme, we preferentially select clients with higher power values by controlling the weights in the utility function. We use different models for the power consumption based on devices' system configuration which accounts for idle and normal usage states of the devices.

\section{Federated Model Training}
\label{sec:training}
In this section, we explain how \scheme quantifies the client utility, how it selects high-utility clients as training unwinds, and then discuss the energy consumption models.

\subsection{Clients utility definition}
The reward function comprises two parts, seeking to reconcile the demand for time-to-accuracy and energy conservation. We modify the current reward function proposed in Oort \cite{Oort-osdi21} by replacing it for client  \textit{i} as:

\begin{equation}
\label{eq:reward}
reward = f  \times  Util(i)  +  (1-f)  \times  power(i),
\end{equation} 
where, $f \in [0,1]$ and \cref{eq:reward} will naturally give high-priority to the high-power clients as $f \rightarrow 0$. For example, to prioritize the clients with high-battery levels we use: \[ power(i) =  cur\_battery\_level(i) - battery\_used(i) \] 

\smartparagraph{Trade-off between system and statistical efficiency: } the first part of the reward function of \cref{eq:reward} influences the time-to-accuracy metric of the FL process. It relies on two factors: (i) \textbf{system efficiency}:  completion time of each training round, and (ii) \textbf{statistical efficiency}: the number of rounds completed to reach the desired accuracy. Both forms of efficiencies are considered jointly for better time-to-accuracy performance. Oort \cite{Oort-osdi21} finds a sweet spot in the trade-off by attaching a utility with every client that optimizes each form of efficiency. In Oort, a high statistical utility may lead to longer rounds especially if the client becomes a bottleneck for the aggregation round; moreover, on the other hand, a high system utility may reduce each round's duration and can lead to more rounds as the fastest clients are exclusively picked who become over-represented. The utility of each client $i$ is formulated by a utility calculated after each training round and is given by \cite{Oort-osdi21}:  

\begin{equation}
\label{eq:oort}
Util(i) = |B_i|\sqrt{ \frac{1}{|B_i|}\sum_{kEB_i}Loss(k^2)} \times (\frac{T}{t_i})^{1(T<t_i) \times a}
\end{equation}

where, in \cref{eq:oort}, $T$ is the duration of each round, $t_i$ is the time taken by the client to process the training, and $1(x)$ is an indicator function that is 1 if $x$ is true and 0 otherwise. More details are given in \cite{Oort-osdi21}.

\begin{table}[!t]
\caption{Comm. energy consumption ($y$) given duration ($x$)}

\begin{tabular}{llcl}
\cline{1-3}
\multicolumn{1}{|l|}{}     & \multicolumn{1}{l|}{\textbf{Download}} & \multicolumn{1}{c|}{\textbf{Upload}}    \\ \cline{1-3}
\multicolumn{1}{|l|}{WIFI} & \multicolumn{1}{l|}{y = 18.09x + 0.17} & \multicolumn{1}{c|}{y = 21.24x - 2.68}  \\ \cline{1-3}
\multicolumn{1}{|l|}{3G}   & \multicolumn{1}{l|}{y = 20.59x - 1.09} & \multicolumn{1}{c|}{y = 15.31x + 2.67}   \\ \cline{1-3}
\end{tabular}

\label{tab:table1}
\end{table}

\subsection{Energy consumption model}

The energy consumption in the second part of \cref{eq:reward} comes from the local computations executed during training and the wireless transmissions of the model updates. %

\smartparagraph{Computation:} The energy consumed by a selected device for a local iteration can be determined by taking the product of execution time and the run-time power such as $E_{comp} = P \times t$ where $t$ is the time spent in the training on CPU/GPU and $P$ are the power consumption at average usage during the training. Here, we assume that the mobile devices are equipped with GPUs and hence we inherit the GPU power model as in \cite{autoflenergy}. The run-time power in the busy state is calculated for each category of edge devices (high-end, mid-end or low-end devices) and is dependent on the device-specific parameters (detail of the device categories and specifications is mentioned in \cref{sec:eval}).

\smartparagraph{Communication:}
The mobile devices participating in a particular round of the FL training transmit their model aggregates to the server via wireless transmissions, corresponding to the communication energy consumption.  We follow a simple linear energy model proposed in \cite{energymodel}. \cref{tab:table1} shows the energy consumption functions for the elapsed time (communication latency) while using WiFi or 3G communication technologies to upload and download data. These functions compute the percentage of battery consumed by smartphones ($y$) when using the respective mediums for $x$ hours. Their measurements were conducted on HTC Desire HD smartphone running Android OS version 2.3~\cite{energymodel}.

\section{Evaluation}
\label{sec:eval}
We evaluate \scheme against Oort and Random selection.

\smartparagraph{Experimental Setup}
In the experiments, we simulate an FL benchmark for speech recognition task that uses the Google Speech dataset~\cite{googlespeech} and ResNet model~\cite{resnet}. Learners are assigned real-world devices and network capability profiles from the AI Benchmark~\cite{AIranking} and MobiPerf Trace~\cite{mobiperf}, respectively. This is an event-driven simulation with time calculated based on the completion time of the learners. To train the model, we use a cluster of GPU servers and interleave the time between individual learners. We assigned 4 GPUs per experiment. We use Fedscale as the training framework~\cite{FedScale} and YoGi as the aggregation algorithm~\cite{YoGi}. The hyper-parameters are set to 0.05, 500, and 20 for learning rate, \# of epochs, and batch size, respectively. For each training round, the target number of learners to be selected is 10, and we make the clients available all the time. For the selected devices, we calculate energy consumption during training using the power model \cref{eq:reward}; however, for unselected devices, we deduce the energy consumed for being in a combination of idle or busy states.  

\smartparagraph{Data Partitioning:} 
The client to data mappings used in Oort~\cite{Oort-osdi21} is close to an IID distribution, so we introduce a more realistic non-IID distribution. The learners are assigned data samples from a random 10\% of the labels (4 out of 35) while the data points per learner are sampled uniformly.

\begin{table}[!t]
\caption{Mobile device specification}

\resizebox{\columnwidth}{!}{
\begin{tabular}{|c|c|c|c|c|}
\hline
\textbf{Device}                                                                       & \textbf{\makecell{Average\\Power (W)}} & \textbf{Perf/W} & \textbf{Memory} & \textbf{\makecell{Battery\\Capacity}} \\ \hline
\textbf{\begin{tabular}[c]{@{}c@{}}Huawei Mate 10 (Kirin 970) \\ (High-end)\end{tabular}}    & 6.33                    & 5.94 fps/W                 & 4GB(RAM)       & 4000mAh                   \\ \hline
\textbf{\begin{tabular}[c]{@{}c@{}}Nexus 6P (Snapdragon 810 v2.1)\\ (Mid-range)\end{tabular}} & 5.44                    & 4.03 fps/W                 & 3GB(RAM)       & 3450mAh                   \\ \hline
\textbf{\begin{tabular}[c]{@{}c@{}}Huawei P9 (Kirin 955)\\ (Low-end)\end{tabular}}          & 2.98                    & 3.55 fps/W                 & 3GB(RAM)        & 3000mAh                   \\ \hline
\end{tabular}
}
\label{tab:table2}
\end{table}

\smartparagraph{Device Profiles:} We evaluate each device's communication and computation profiles using the real device measurements from AI benchmark ~\cite{AIranking} and MobiPerf~\cite{mobiperf} benchmark. However, for energy consumption, there is no such information widely available hence we cluster these profiles into three main performance categories (high, mid, and low) and map each device to one of them. We select three smartphones to represent the high, mid and low-end categories. Their average power consumption measurements from GFXBench \cite{gfx} along with device specifications are shown in \cref{tab:table2}. %

 \begin{figure*}[!t]
    \begin{subfigure}{0.33\linewidth}
        \includegraphics[width=\linewidth]{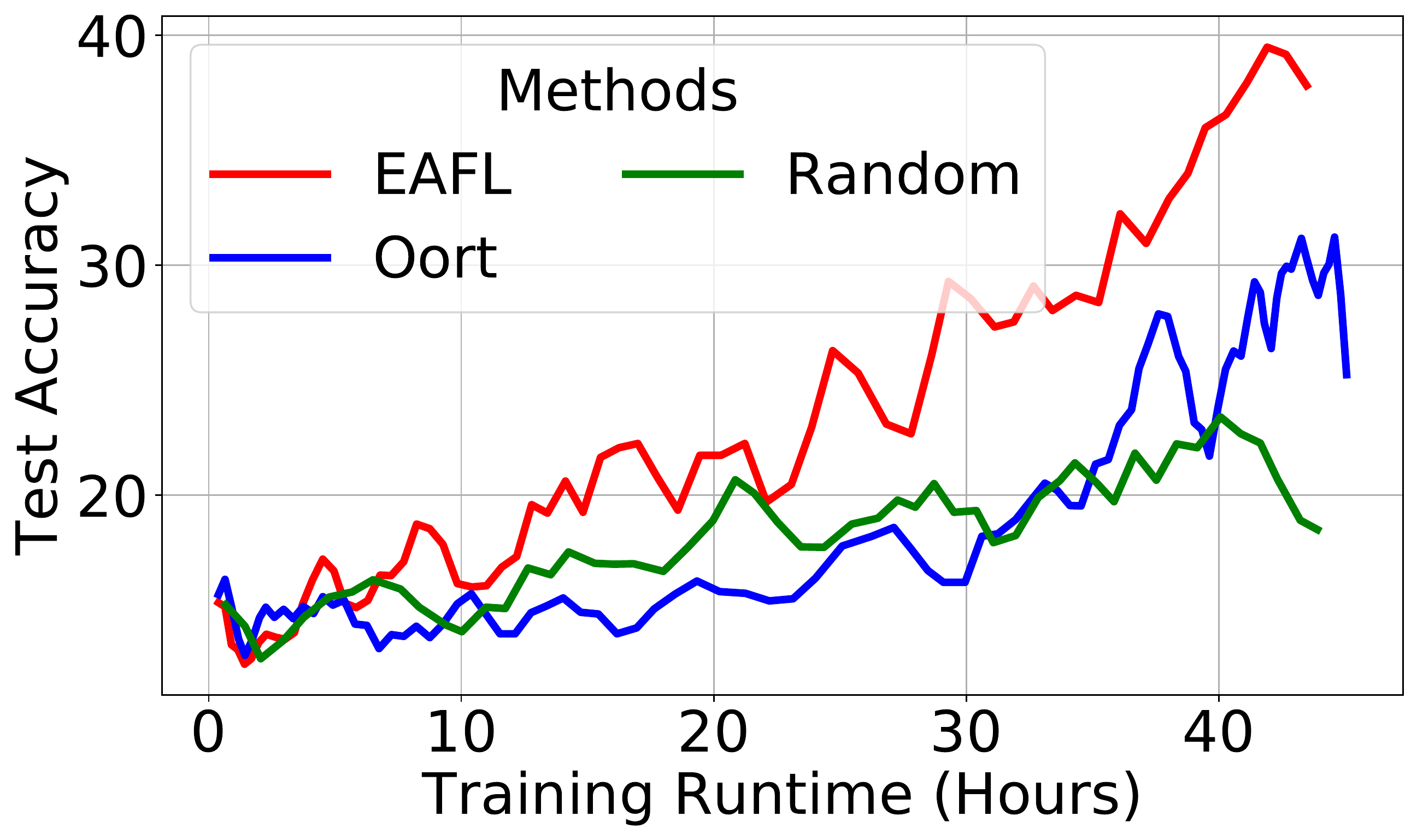}
         \caption{Test accuracy vs time}
         \label{fig:accuracy}
    \end{subfigure} 
    \hfill
    \begin{subfigure}{0.33\linewidth}
        \includegraphics[width=\linewidth]{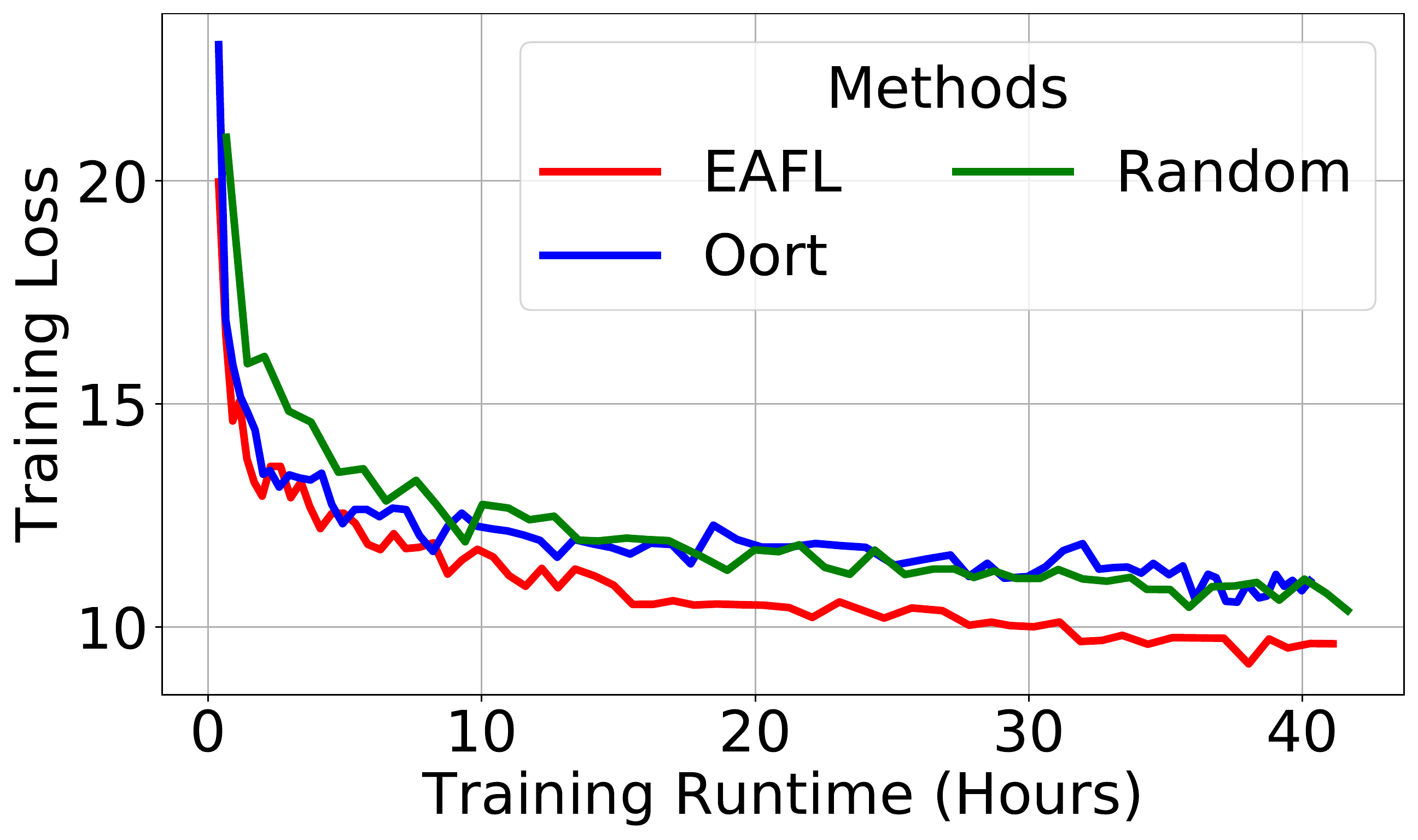}
        \caption{Training loss vs time}
         \label{fig:loss}
    \end{subfigure}
    \hfill
    \begin{subfigure}{0.33\linewidth}
        \includegraphics[width=\linewidth]{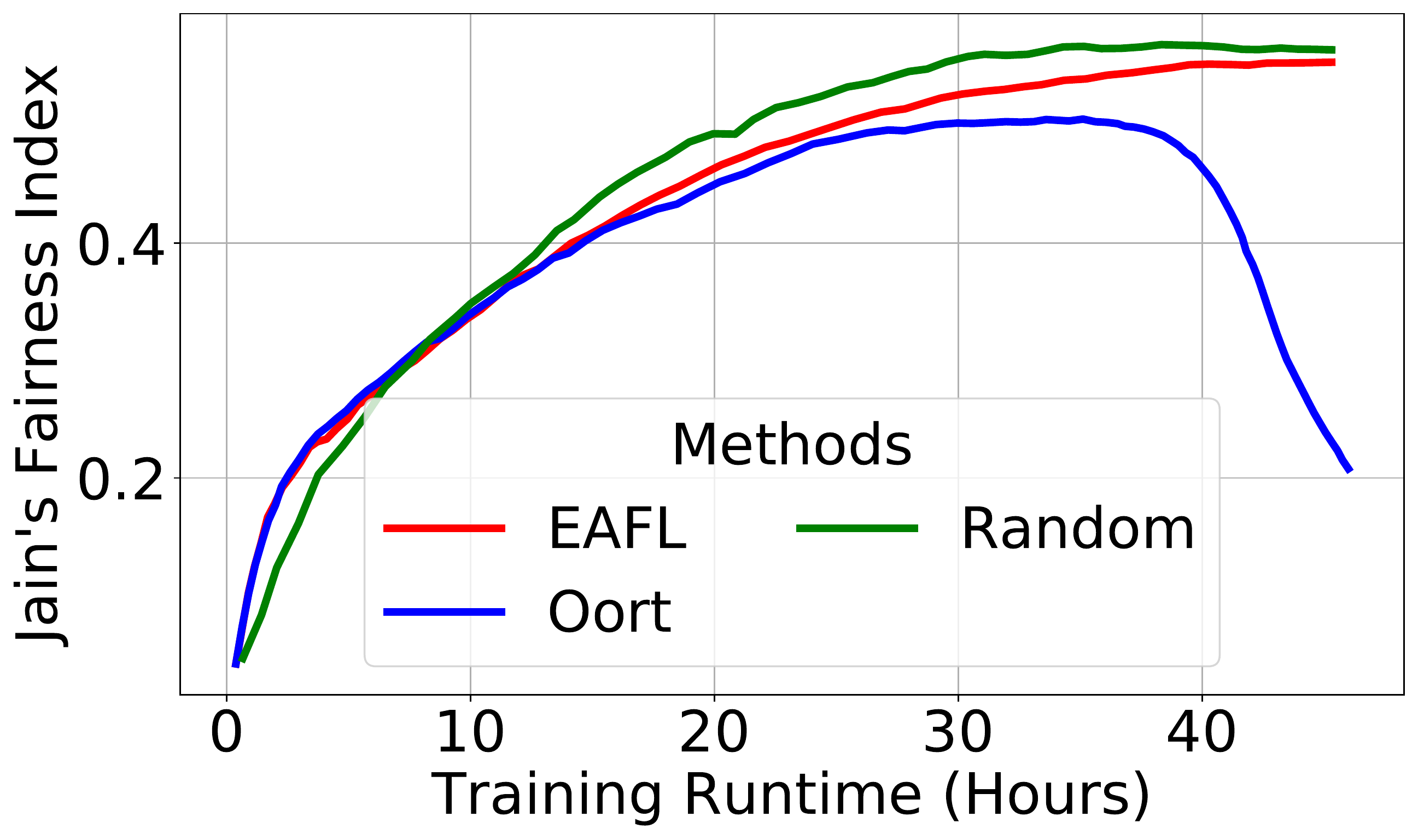}
         \caption{Jain's fairness index vs time}
         \label{fig:fair}
    \end{subfigure} 
  \label{fig:results}
  \caption{Performance of Oort and \scheme in terms of (a) test accuracy, (b) train loss, (c) Jain's fairness index in a Non-IID case.}

\end{figure*}

\begin{figure}[!t]
 \begin{subfigure}{0.48\linewidth}
    \includegraphics[width=1\linewidth]{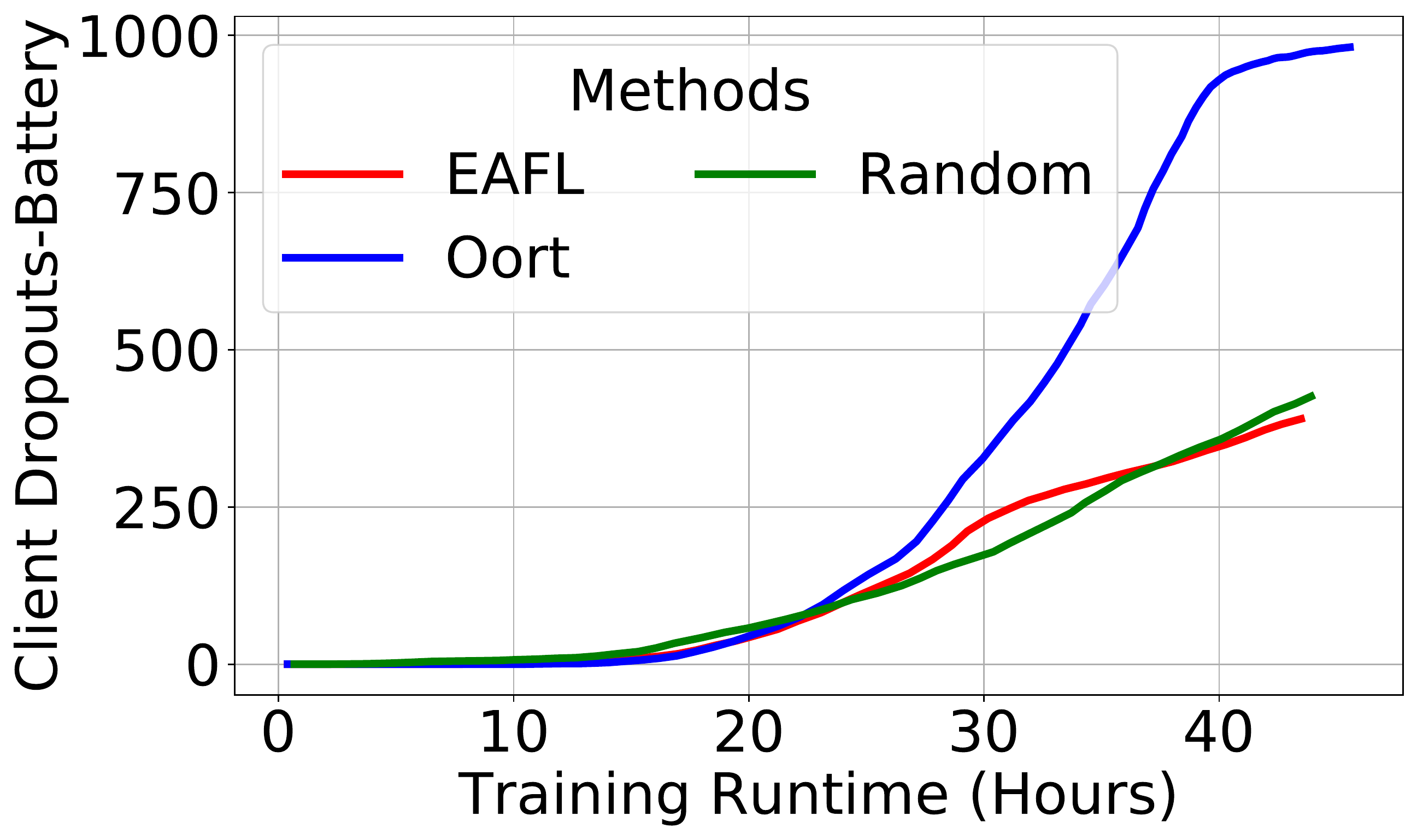}
  \caption{Clients running out of battery}
  \label{fig:dropouts}
  \end{subfigure}
  \hfill
  \begin{subfigure}{0.48\linewidth}
  \includegraphics[width=1\linewidth]{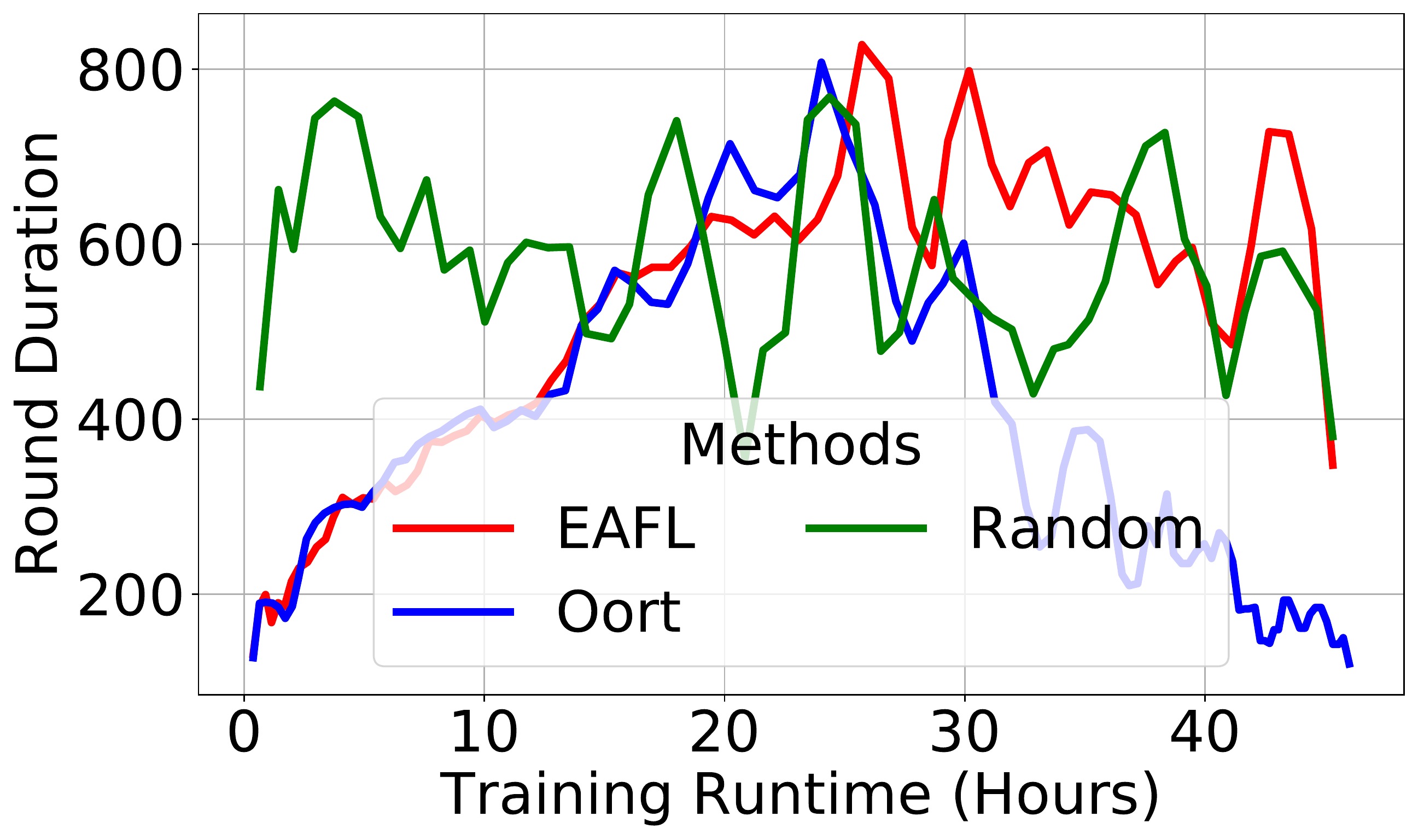}
    \caption{Round duration}
  \label{fig:duration}
  \end{subfigure}
  
  \caption{Comparison of client dropouts due to running out of Battery between Oort and \scheme.}
  \label{fig:metrics}
\end{figure}

\smartparagraph{Experimental Results:} To evaluate the performance of \scheme, we compare it with Oort and a random sampler (Random). We run the experiments in non-iid settings. We use the value of $f=0.25$ in \cref{eq:reward} to calculate the reward function for \scheme to give more weight to the high-power clients (i.e., higher remaining battery). This allows more clients to stay and reduces dropouts due to running out of battery. 

As shown \cref{fig:dropouts}, compared to other methods, Oort shows a significant increase in battery run-outs as the training progresses, highlighting the impact of not being energy-aware unlike \scheme and random with low drop-outs. \scheme, to conserve energy, it trades-off time and selects high-power clients and hence sees fewer drop-outs over the training. Random has low dropouts as it selects clients uniformly at random because it distributes the consumption over more clients, however, its consumption supersedes \scheme after 40 hours of training. \cref{fig:duration} shows Random results in significantly higher round duration compared to the other methods. We note the per-round duration for Oort and \scheme is almost the same, but further in time, the duration increases to compensate for low statistical efficiency (Oort) or high energy consumption (\scheme). Oort experiences a low participation rate and round failures due to significant numbers of dropouts which explains its low-round duration at the end. \scheme is able to maintain low client drop-outs due to its energy-aware selection which increases the participation rate in the training. %

\cref{fig:accuracy} and \cref{fig:loss} present the achieved accuracy and training loss for all methods, respectively. \scheme achieves the best results in both metrics showing the benefits of its selection method (esp. for battery-powered scenarios) which maintains high levels of participation in the training. \cref{fig:fair} shows Jain’s fairness index for device selection which measures if users are getting a fair opportunity to participate in the training. The results suggest that \scheme enjoys high levels of fairness similar to Random. Oort initially enjoys the same levels of fairness but then due to high dropouts and low participation, its fairness degrades severely. 

\section{Related Work}
\smartparagraph{Federated Learning (FL):} is a new distributed machine learning method which is increasingly becoming popular for its privacy-preservation and low-communication features. This motivated the growing adoption of FL to improve the end-user experience (e.g., the search suggestion quality of virtual keyboards~\cite{yang2018applied}). Moreover, to encourage and speed up experimentation of new ideas, many FL frameworks were recently developed~\cite{caldas2018leaf,ryffel2018generic}. In FL, the training of a global model is assigned to a sub-population of decentralized devices such as mobile or IoT sensor devices. These devices possess private data samples and engage in training the model on their local data~\cite{li2014scaling,mcmahan2017}.   

\smartparagraph{System heterogeneity:} One of the major contributors to system performance unpredictability is the heterogeneity inherent in many of the distributed systems. Mainly, in FL context, the heterogeneity of devices' system configurations (e.g., computation, communication, battery, etc) results in unpredictable performance. For instance, the stragglers (i.e., slow workers) can halt the training process for a prolonged duration~\cite{Li2020FedProx,Ahmed-AQFL-21}. 
Several solutions exist that address this problem through system and algorithmic solutions~\cite{Ahmed-IPCCC-2015,Ahmed-ICC-2016-2,Li2020FedProx,Yichen2021,Ahmed-AQFL-21}. Moreover, in FL, the heterogeneity is also a byproduct of other artifacts other than the devices. For example, the learner data distributions, the participants' selection method, and the behaviour of the device's owner are common sources of heterogeneity in FL setting~\cite{Ahmed-FLSurvey-2021,Ahmed-EuroMLSys-22}. %

\smartparagraph{Energy-conservation:} Considering the uncertainties in the mobile environment, several energy management techniques have been proposed~\cite{autoflenergy}. %
Other works for energy efficiency optimization in the mobile environment use computation offloading techniques. Finally, few works focused on energy-efficient FL training~\cite{autofl,autoscale}. \scheme aims for energy conservation to mitigate client dropouts which is the major contributor to the degraded FL model qualities.

\smartparagraph{Improvements in FL:} 
In FL, several works aim to improve the time-to-accuracy of training by leveraging techniques such as periodic updates, compression, and layer-wise asynchronous updates
\cite{Jakub2016,Bonawitz19,Ahmed-AAAI-2020,pmlr-v108-reisizadeh20a,Ahmed-DC2-INFOCOM21,Ahmed-RELAY-2021,Ahmed-SIDCo-MLSys21,Ahmed-huffman-2020,grace-2021,Ahmed-NeurIPS-2021}. Other proposals aimed at righting the privacy guarantees of FL settings \cite{Melis2019,Bonawitz19,Nasr2019,Bagdasaryan20}. %
Moreover, the bias in FL is studied to ensure fair participation in the training  process~\cite{Mohri2019AgnosticFL,Li2020FedProx}. 

\section{Conclusion}
\label{sec:conclusion}

We present our preliminary work on studying a practical federated learning scenario where the participants are battery-powered and they need to collaboratively learn a new global model. We find that the existing state-of-the-art methods, aiming to minimize the time-to-accuracy, fail to achieve satisfactory performance as they ignore the factor of energy consumption on the devices. Therefore, we present \scheme, to enable power-aware FL training on battery-powered devices. Our algorithm intelligently selects the participants to maintain low time-to-accuracy while conserving energy. It shows up to 85\% improvement in model accuracy and 2.45$\times$ decrease in drop-out of clients proving to be a practical FL solution for battery-powered scenarios.

\footnotesize
\bibliography{references,mypapers}
\bibliographystyle{ACM-Reference-Format}
\end{document}